\documentclass[]{spie}  
\usepackage[]{graphicx}

\usepackage{tikz}
\usetikzlibrary{positioning}
\usetikzlibrary{shapes,arrows,chains}
\usetikzlibrary{shapes.geometric}
\usepackage{amssymb, amsmath}
\DeclareMathOperator{\Ima}{Im}
\usepackage{array, float}
\usepackage{url}

\usepackage{ragged2e}
\newcolumntype{P}[1]{>{\RaggedRight\hspace{0pt}}p{#1}}

\usepackage[colorinlistoftodos,prependcaption,textsize=small,textwidth=3.5cm]{todonotes}

\usepackage[abs]{overpic} 

\newcommand{\etal}{{et al}.\@ }

\usepackage[acronym]{glossaries}

\newacronym{ADDA}{ADDA}{adversarial discriminative domain adaptation}
\newacronym{AiTR}{AiTR}{aided target recognition}
\newacronym{ATR}{ATR}{automatic target recognition}

\newacronym{BD}{BD}{Bregman Divergence}
\newacronym{BKR}{BKR}{Biomedical Knowledge Repository}
\newacronym{BRCA}{BRCA}{Breast Invasive Carcinoma}

\newacronym{CNN}{CNN}{convolutional neural network}
\newacronym{CoGAN}{CoGAN}{coupled generative adversarial nets}

\newacronym{CyCADA}{CyCADA}{cycle-consistent adversarial domain adaptation}

\newacronym{DA}{DA}{domain adaptation}
\newacronym{DAd}{DAd}{domain adversarial}

\newacronym{DAN}{DAN}{deep adaptation network}
\newacronym{DiSDAT}{DiSDAT}{direct sum domain adversarial transfer}
\newacronym{DL}{DL}{deep learning}

\newacronym{DM}{DM}{diffusion map}

\newacronym{DNN}{DNN}{deep neural network}

\newacronym{DSDA}{DSDA}{direct sum domain adaptation}
\newacronym{DS}{DS}{direct sum}

\newacronym{EMA}{EMA}{exponential moving average}

\newacronym{EO}{EO}{electro-optical}

\newacronym{FC}{FC}{fully-connected}

\newacronym{GAN}{GAN}{generative adversarial network}
\newacronym{GB}{GB}{gigabyte}
\newacronym{GPU}{GPU}{graphical processing unit}
\newacronym{GCN}{GCN}{graph convolutional network}

\newacronym{HHTL}{HHTL}{Hybrid Heterogeneous Transfer Learning}
\newacronym{HTL}{HTL}{heterogeneous transfer learning}

\newacronym{JAN}{JAN}{joint adaptation network}
\newacronym{JMMD}{JMMD}{joint maximum mean discrepancy}

\newacronym{KDE}{KDE}{kernel density estimation}

\newacronym{kNN}{k-NN}{k-nearest neighbor}
\newacronym{KL}{KL}{Kullback–Leibler}

\newacronym{LUAD}{LUAD}{Lung Adenocarcinoma}
\newacronym{MacL}{ML}{machine learning}
\newacronym{MMD}{MMD}{maximum mean discrepancy}
\newacronym{MC}{MC}{Monte Carlo}

\newacronym{MR.Net}{MR.Net}{Manifold Regularized Network}

\newacronym{mSDA}{mSDA}{marginalized stacked denoised autoencoder}
\newacronym{MSE}{MSE}{mean square error}

\newacronym{NLM}{NLM}{National Library of Medicine}
\newacronym{NN}{NN}{neural network}

\newacronym{RAM}{RAM}{random access memory}

\newacronym{RCA}{RCA}{Regularized Conditional Alignment}

\newacronym{ReLU}{ReLU}{rectified linear units}
\newacronym{RKHS}{RKHS}{reproducing kernel Hilbert space}

\newacronym{SAR}{SAR}{synthetic aperture radar}
\newacronym{SGD}{SGD}{stochastic gradient descent}

\newacronym{SoA}{SoA}{state-of-the-art}
\newacronym{SVM}{SVM}{support vector machine}

\newacronym{TDA}{TDA}{topological data analysis}

\newacronym{TrDM}{TrDM}{transfer diffusion map}

\newacronym{TL}{TL}{transfer learning}
\newacronym{TLDA}{TLDA}{transfer learning with deep autoencoders}
\newacronym{TSL}{TSL}{transfer subspace learning}

\newacronym{VADA}{VADA}{Virtual Adversarial Domain Adaptation}

\newacronym{VAT}{VAT}{virtual adversarial training}

\makeglossaries

\title{Domain Adaptation by Topology Regularization}

\author{Deborah Weeks\supit{a} and Samuel Rivera\supit{b}
\skiplinehalf
\supit{a}George Washington University, Washington DC, USA;\\
\supit{b}Matrix Research, Dayton, USA   
}

\authorinfo{Further author information: (Send correspondence to S.R.)\\
S.R.: E-mail: samuel.rivera@matrixresearch.com \\ 
}
\begin{document}
  \maketitle

\begin{abstract}

Deep learning has become the leading approach to assisted target recognition. While these methods typically require large amounts of labeled training data, \gls{DA} or \gls{TL} enables these algorithms to transfer knowledge from a labelled (source) data set to an unlabelled but related (target) data set of interest. \gls{DA} enables networks to overcome the distribution mismatch between the source and target that leads to poor generalization in the target domain. \gls{DA} techniques align these distributions by minimizing a divergence measurement between source and target, making the transfer of knowledge from source to target possible. While these algorithms have advanced significantly in recent years, most do not explicitly leverage global data manifold structure in aligning the source and target. We propose to leverage global data structure by applying a \gls{TDA} technique called persistent homology to \gls{TL}.

In this paper, we examine the use of persistent homology in a  \gls{DAd} \gls{CNN} architecture. The experiments show that aligning persistence alone is insufficient for transfer, but must be considered along with the lifetimes of the topological singularities. In addition, we found that longer lifetimes indicate robust discriminative features and more favorable structure in data. We found that existing divergence minimization based approaches to \gls{DA} improve the topological structure, as indicated over a baseline without these regularization techniques. We hope these experiments highlight how topological structure can be leveraged to boost performance in \gls{TL} tasks.

\end{abstract}




\section{Introduction}
\label{sec:intro}  
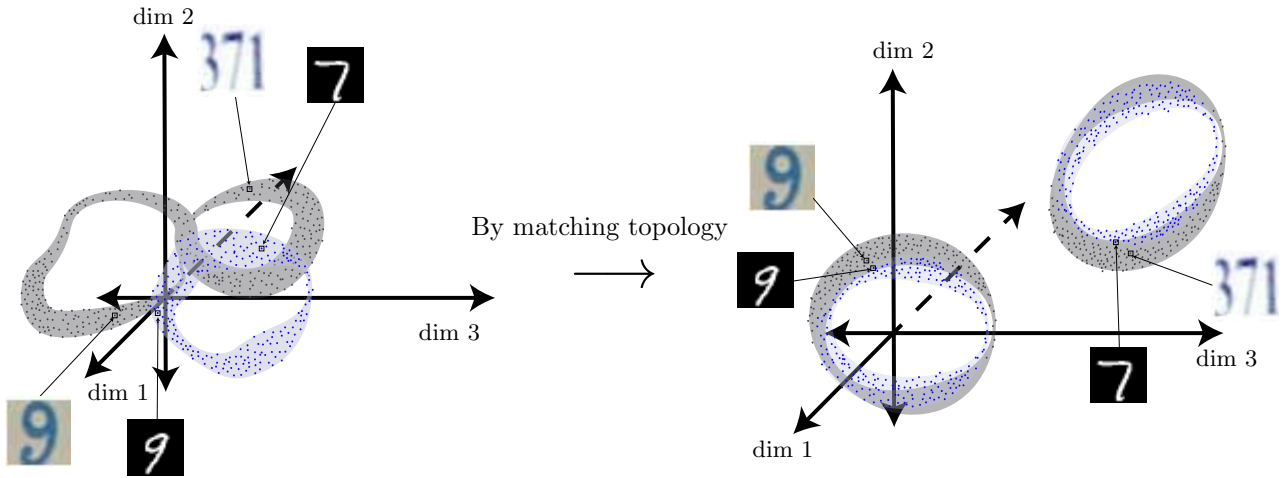
\begin{figure}[ht]
  \begin{centering}
  $ \vcenter{\hbox{\begin{overpic}[unit=1mm, scale = .75]{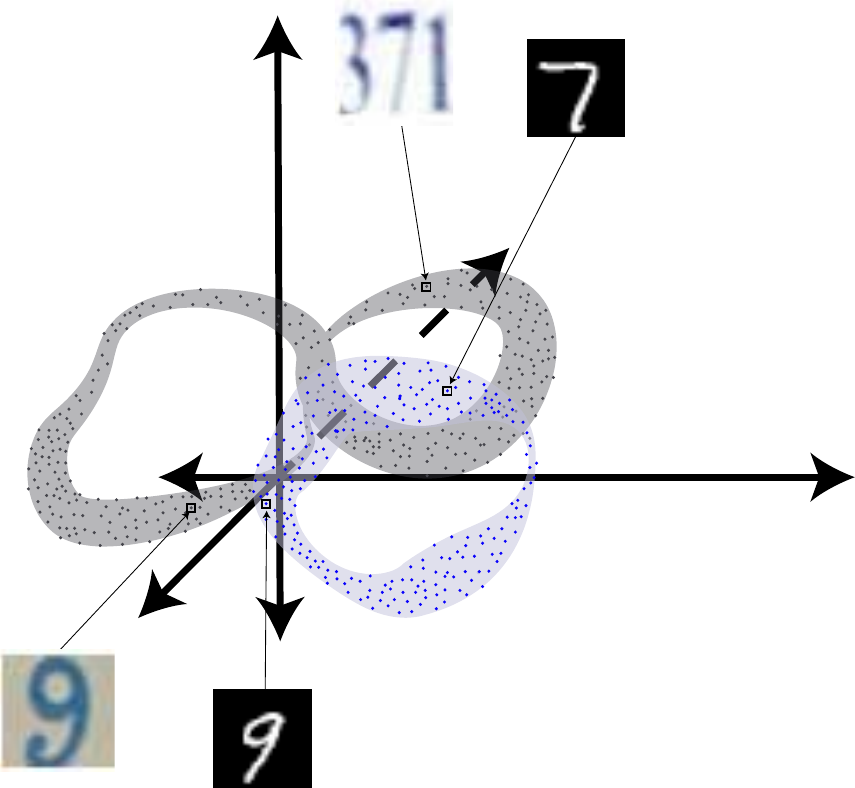}
  \put(11, 10){\small{dim 1}}
  \put(17,60){\small{dim 2}}
  \put(55, 18){\small{dim 3}}
  \put(75,25){\huge{$\longrightarrow$}}
  \put(62,32){By matching topology}
  \end{overpic}}}$ 
  \hspace{3cm}
  $\vcenter{\hbox{\begin{overpic}[unit=1mm, scale = .75]{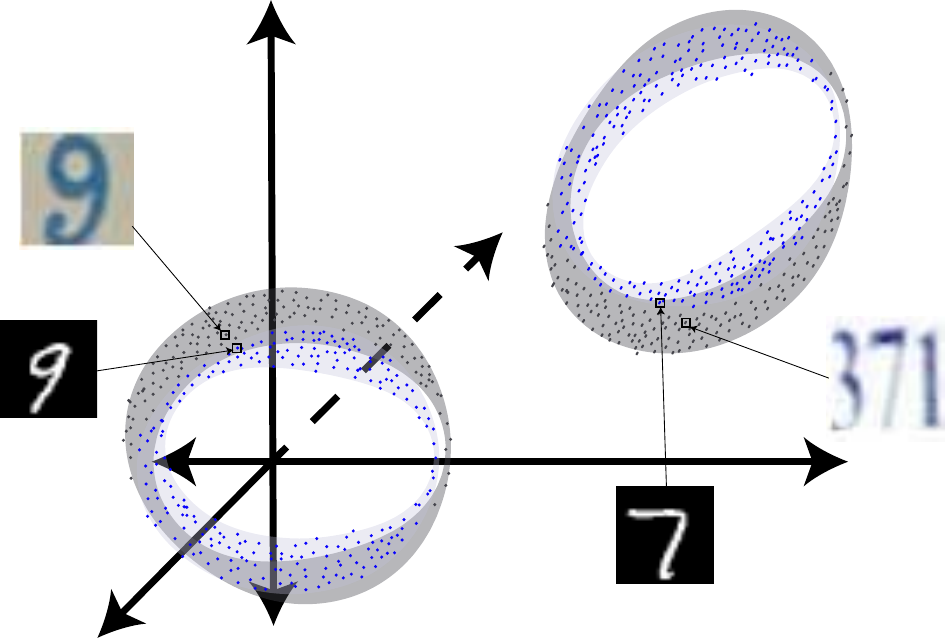}
  \put(2, -3){\small{dim 1}}
  \put(18,50){\small{dim 2}}
  \put(61,9){\small{dim 3}}
  \end{overpic}}}$
    \end{centering}
        \caption{Manifolds represent global data structure. Here the grey disk with two holes represents the source data manifold, while the dark grey dots represent individual source image examples. The blue annulus represents the target data manifold, and the dark blue dots represent the target image examples. The left side image shows a possible outcome of feature extraction without topological considerations. The right side shows the theoretical outcome with topology  regularization.}
        \label{introfigure}
    \end{figure}

Recent advances in the field of machine learning have focused largely on deep neural networks. These networks learn feature representations that can be used in a variety of applications including object recognition, natural language processing, and scene segmentation from sensor data.  Unfortunately, deep neural networks need large quantities of labeled training data. Moreover, the data must be domain and task specific. For example, classifying images of vehicles requires training images taken during the day, night, and under a variety of weather conditions for the system to work reliably under those various conditions. Unfortunately this type of data can be expensive or impractical to obtain. This lack of labelled data could be resolved if deep neural networks could be trained on one data set, then \emph{transferred} to a different data set. Unfortunately, different dataset features may be distorted such that the probability distributions do not match, which leads to poor generalization \cite{Pan2010}. 

Mitigating the representation shift is a primary focus of \gls{DA} research. Previous attempts fine tune networks to the new data set \cite{Yos}. This approach assumes the source and target feature spaces are similar and does not accommodate for the large differences in data distributions that can come from different sensor types. \gls{DA} attempts to overcome this by aligning the feature representation of a labeled data set - called the source data - with that of an unlabeled data set - called the target data. Past attempts have aligned the distributions by minimizing a measure of divergence  - such as by maximum mean discrepency \cite{Long2017, Long2018}, Kullbach-Leibler divergence\cite{Kullback51}, or a domain loss\cite{ADDA}. Some of these measures emphasize the local interactions between points. By contrast, a topological regularization would use the global structure of the source and target data manifolds  to align them in the common latent space.

In this paper we investigate using persistent homology, a tool from \gls{TDA}, to analyze and align the global structure of the source and target feature representations of a \gls{NN}. The idea is illustrated in Fig.~\ref{introfigure} for source and target datasets. This method first constructs a manifold from a point cloud by connecting data points that are within some variable threshold distance. Persistent homology defines a persistence diagram by summarizing the number and size (lifetimes) of holes and voids (topological singularities) in the data manifold formed by the point cloud. These persistence diagrams summarize the topological structure, and can be used to regularize the network's feature representation.

Our empirical findings are summarized as follows: We examined what features persistent homology can detect in data samples. We found that longer lifetimes indicate favorable features and structure in the data manifold. We then analyzed what impact current state of the art techniques have on the global structure of feature representations and where in the network architecture this structure can be useful. Existing divergence minimization based approaches to \gls{DA} improved the topological structure, as indicated by longer lifetimes over a baseline network. Also, we found that as data goes through layers of \gls{CNN}, there is an increase in topological structure after passing through the second convolutional layer. This increase is later broken down toward the discriminative parts of the network architecture. Guided by the results of these experiments, we designed a network that forces the global structure of the source and target feature representations to be topologically similar by minimizing the Wasserstein distance between their persistence diagrams in the latent network layers. We found no transfer improvement beyond the existing \gls{DA} approaches, and discuss possibilties for this finding. Taken collectively, our experiments show that persistent homology is useful for identifying robust representations in general, but may be less useful for divergence minimization.

The paper is organized as follows: We provide the technical background in Sec.~\ref{sec:Background} before reviewing related work and the methodology in Sections~\ref{sec:related} and \ref{sec:methods}, respectively. The experiments and analysis of Sec.~\ref{sec:experiments} provide the bulk of our reserach contribution. Finally, we close with a discussion and conclusion.
\section{Background}
\label{sec:Background}

Our work uses \gls{TDA} to examine the manifold structure of the feature representations given by a \gls{TL} network. \gls{TDA} uses tools from algebraic topology to better understand the structure of data manifolds.  The goal of this section is to define homology and explore its application to point cloud data. This background is necessary in order to fully understand the differentiation of the topological layers of our network, and the reasoning behind the specific choices made in regularizing the source and target features with respect to topology. For a more in depth look at algebraic topology, see the references by Hatcher \cite{Hatcher} and Munkres \cite{Munkres}. We follow these sources in our definitions of the main terms. In the second part of this section, we introduce homology, persistence, and current theoretical results that are pertinent to machine learning. In our explanation of persistent homology, we follow the surveys by Edelsbrunner and Harer \cite{phsurvey} and Ghrist \cite{Ghrist}. More details can be found in the text by Edelsbrunner and Harer \cite{computetopbook}.

We first relate homology theory \cite{HomologyOrigins} to the study of deep neural network representations. Feature representations can be subspaces of high dimensional vector spaces. It can be difficult to detect the presence of a gap or void in such a high dimensional subspace because they are difficult to visualize. Homology reveals features in the point cloud that would otherwise be invisible to researchers \cite{Ghrist}. Generally, homology measures how connected a manifold is. It can be thought of as counting the number of holes, gaps or voids in the manifold structure. This is important for \gls{TL} because ideally feature representations  form tight clusters with individual clusters housing a single class \cite{clusterassumption}. 

Our network constructs a manifold that approximates a feature representation and analyzes its clustering behavior. This construction uses the most basic objects in algebraic topology: simplices, simplicial complexes, chain groups and chain complexes. These definitions naturally build on each other from simplices up to chain complexes. A \textbf{simplex} (plural simplices) is a generalization of a triangle or tetrahedron to arbitary dimensions \cite{Hatcher}.  A $0$-simplex would be a single point, a $1$-simplex would be a line segment, a $2$-simplex would be a solid triangle, and a $3$-simplex would be a solid tetrahedron. 

To see how homology reveals structure in point clouds, we first must make these definitions mathematically precise. Formally, the standard $n$-simplex with vertices that are the unit vectors in $\mathbb{R}^n$ is the set:
 $$\Delta^n := \{ (t_0,...,t_n) \in \mathbb{R}^{n+1} | \sum_i t_i = 1 \ and \ t_i \geq 0 \ \forall \ i \}.$$
 
We can shift this standard $n$-simplex by using different vertices, $v_0,..., v_n$, and define the more general simplex as the smallest convex set in a Euclidean space $\mathbb{R}^m$ containing the $n+1$ points $[v_0,...,v_n]$ with the additional requirement that the difference vectors $[v_1 - v_0, ..., v_n - v_0]$ are linearly independent \cite{Hatcher}. If one vertex of the $n$-simplex is deleted then the remaining vertices span an $(n-1)$-simplex called a face of the original simplex. Intuitively, these would be an endpoint of a line segment, an edge of a triangle, or a face of a tetrahedron. The boundary of a simplex consists of all of the faces of the simplex.  So, the boundary of a line segment (1-simplex) would be the two endpoints. A \textbf{simplicial complex} is a collection of simplices that have been glued together along their faces according to some plan \cite{Munkres}. For example, a hollow triangle could be represented as a simplicial complex by using three  $0$-simplices, $\{a,b,c\}$ and three $1$-simplices, $\{ \overrightarrow{ab}, \overrightarrow{bc}, \overrightarrow{ca} \}$, where $\overrightarrow{ab}$ denotes the line segment connecting point $a$ to point $b$, see Figure \ref{triangle}: 

 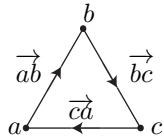
\begin{figure}[ht]
  \begin{centering}
$ \vcenter{\hbox{\begin{overpic}[unit=1mm, scale = .5]{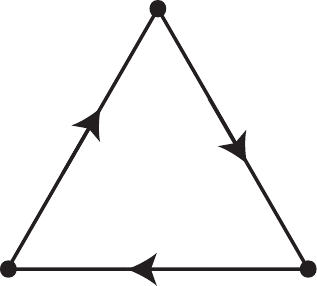}
\put(-2,0){$a$}
\put(8,15){$b$}
\put(14, 7){$\overrightarrow{bc}$}
\put(17,0){$c$}
\put(6,2){$\overrightarrow{ca}$}
\put(-1,7){$\overrightarrow{ab}$}

\end{overpic}}}$

  \end{centering}

      \caption{
     A triangular simplicial complex. Note that this is only the exterior part of the triangle.}
      \label{triangle}
  \end{figure}

Recall that our motivation is to analyze the gaps and voids of feature representations using homology theory. After formally defining homology, we will explain how homology accomplishes this, and how backpropogation can be implemented across these measurements. To formally define homology, we must turn to the theory of abstract algebra. We place an algebraic structure on the collection of simplices found in a simplical complex.  The \textbf{chain group}, $C_k$, is the set of linear combinations of simplices of dimension $k$ in the simplicial complex $C$. Let $\overrightarrow{xy}$ denote the directed line segment connecting point $x$ to point $y$. So for our triangle example, $b-a \in C_0$ and $\overrightarrow{ab} + \overrightarrow{bc} - \overrightarrow{ca} \in C_1$. In \gls{TDA}, we take the linear combinations over $\mathbb{Z}/2\mathbb{Z}$, where  $\mathbb{Z}/2\mathbb{Z}$ denotes the finite field with two elements - $\{0,1\}$ \cite{phsurvey}.

In addition to the gluing directions, \textbf{chain complexes} also include a boundary map $\partial: C_k \rightarrow C_{k-1}$. Intuitively speaking, the boundary map takes a simplex in $C_k$ and maps it to its boundary which is represented in $C_{k-1}$. In our triangle example above, we could define the boundary map $\partial$ so that it takes $\overrightarrow{ab}$ to $b-a \equiv b+a $, where here the equivalence ($\equiv$) is in $\mathbb{Z}/2\mathbb{Z}$. These boundary maps have the property that $\partial_{k-1} \circ \partial_{k} = 0$, where $f \circ f$ denotes the composition of functions: $f(g(\cdot))$.

We now have the necessary background to define homology and understand its application to point cloud data. A standard algebraic approach for studying large chain groups is to instead examine smaller subchains. Subchains retain the same operators, but only look at a subset of group elements that satisfy a specific property. The boundary maps associated with the chain groups $C_k(M)$ and $C_{k+1}(M)$ can be used to define two subchains. The $k$-chains (elements of $C_k(M)$) that have boundary $0$ are called $k$-cycles. They are sometimes also referred to as the kernel of the map $\partial_k$, $\text{ker}  \ \partial_k$. This kernel is the same as the null-space from linear algebra. The $k$-chains that are a boundary of $(k+1)$ chains are called $k$-boundaries. They are the image of the $k+1$ level boundary map, denoted $\Ima \partial_{k+1}$. This is analogous to the span of a map from linear algebra.  Because $\partial_{k+1} \circ \partial_k = 0$, $\text{ker} \ \partial_k \subset \Ima \partial_{k+1}$. 

We can begin to see at this point how voids in the data manifold may be analyzed using these sub-chains. The $p$-cycles can be thought of as loops traversing the edges of the simplicial complex. In terms of our point cloud, these are loops that can be travelled on the data manifold without departing it. We wish to consider only the loops that surround empty spaces, since these loops will capture the gaps and voids in the manifold. These are loops without centers. More precisely, they are the loops that are not $k$-boundaries, that is not in $\Ima \partial_{k+1}$. A standard technique from algebra is to examine what properties of a group can be learned by considering as equivalent all elements within a subset of that space. This process is called taking the quotient, and can reveal structure in the underlying space that is not otherwise apparent. The kernel, $\text{ker} \ \partial_k$, and the image, $\Ima \partial_{k+1}$, subgroups can be used to define a quotient by considering equivalent all elements of the image. In terms of our loops, this means any loop that surrounds part of the manifold (i.e. a filled in loop) is equivalent to $0$, and we only consider linear combinations of those loops that are surrounding empty space. The resulting quotient is the homology group. We denote this group by $H_k(M) = \text{ker} \ \partial_k/ \Ima \partial_{k+1}$. We call $k$ the level of the homology. A basis element will appear for each void in the data manifold, and in this way the homology captures the porosity of the space. The dimension of $H_k(M)$, called the Betti number, counts the number of $k$-dimensional topological singularities. For $k=0$, this dimension gives the number of connected components (clusters). For $k=1$, this gives the number of voids or tunnels. For $k=2$, it gives the number of three dimensional balls that have been removed, and so on for higher dimensions.  

Homology is an invariant of a topological space because any continuous deformation of the space does not change its homology type, and will not change the Betti number. It is a way of encoding information about how connected or porous the manifold is.  The cluster assumption of machine learning is fundamentally an assumption about the $0$-level homology of the data manifold of the feature representation. Unfortunately, homology is a theoretical construct. In order to compute the homology of a manifold, we must know its structure from the simplicial complex. In feature representations, we do not know this a priori. However, \gls{TDA} attempts to bridge this gap by using an extension of homology theory - called persistent homology \cite{persistence}. Persistent homology applies the ideas of homology to actual data sets. In persistent homology, a manifold is constructed from a point cloud by connecting points that are within some threshold distance $\alpha$. It analyzes how the homology representation changes as the threshold varies. We now define persistent homology formally.

A naive way to construct the points in the feature representation into a manifold object is to use the point cloud as vertices and to connect edges within some distance. This forms a combinatorial graph. This structure would capture the clustering behavior necessary to analyze $H_0$, but would be unable to detect higher order features. To form the simplicial complex we choose a method for filling in the higher dimensional simplices of the graph. The method chosen in our work is the Rips Complex \cite{Rips}. This was chosen because it is less computationally expensive then other methods \cite{Ghrist}.

Two choices define a Rips Complex - a threshold value, $\epsilon$, and a distance measure, usually Euclidean distance. It is formed by creating a simplex for every finite set of points that has diameter at most $\epsilon$.  The $0$-simplices in this complex are therefore the individual data points. The $1$-simplices are the edges created by connecting the points within $\epsilon_i$ by an edge. The $2$-simplices are filled in triangles formed by three points within $\epsilon$ of each other, and so on for higher dimensions. Formally, given a collection of sample data points $\{x_0, x_1, x_2,..., x_n \}$ in $\mathbb{R}^n$, the Rips Complex is the simplicial complex where each $k$-simplex is an unordered $(k+1)$ tuple of points, $\{ v_0,...,v_{k+1}\}$ such that for all $i \neq j, ||v_i - v_j|| < \epsilon$ \cite{Ghrist}.

Now, if only one value of $\epsilon$ is considered, the $k^{th}$ homology of the Rips Complex gives the number of $k$-dimensional holes for that particular $\epsilon$. This is not enough. Some holes are very small, and can safely be ignored, but some are larger and more essential to the structure of the point cloud. Homology does not differentiate between these two situations. We therefore require more then just one Rips Complex, and instead take a filtration \cite{Ghrist}. A filtration is a hierarchical sequence of Rips complexes:  $X_1 \subset X_2 \subset ... \subset X_n$ formed by increasing the distance $\epsilon_i$ and creating a new Rips Complex, $X_i$. The persistent homology is the collection of homology groups formed when the homology is computed for each $X_i$ in the filtration.  

When a new simplex is added to the complex a topological singularity is created or destroyed \cite{Gabrielsson2019_topolayer, Poulenard}. New simplices are added when the parameter $\epsilon$ is increased. The birth time ($b$) of a singularity is the value of $\epsilon$ where the feature is created. The death time ($d$) is the value corresponding to when it is destroyed.  The output of the persistent homology computation is called a persistence diagram. It consists of an unordered list of $(b,d)$ pairs for each topological singularity present in some $X_i$ \cite{Gabrielsson2019_topolayer}.

The fact that each singularity can be linked to a particular simplex allows for the creation of a differentiable inverse map from the persistence diagrams back to the point cloud \cite{Poulenard}. We can compute the derivative of a functional, $\mathcal{E}$ defined on the persistence diagram \cite{SRpointcloud, Gabrielsson2019_topolayer}:

$$ \frac{\partial \mathcal{E}}{\partial \sigma} = \sum \limits _{i \in I_k} \frac{\partial \mathcal{E}}{\partial b_i} \mathbb{I}_{\pi_f(k)(b_i) = \sigma} + \sum \limits_{i \in I_k} \frac{\partial \mathcal{E}}{\partial d_i} \mathbb{I}_{\pi_f(k)(d_i) = \sigma}$$
Here, the map $\pi_f(k): \{b_i, d_i\}_{i\in I_k} \rightarrow (\sigma, \tau)$, is the inverse map from each $(b_i,d_i)$ pair in the persistence diagram back to the simplex $(\sigma_i, \tau_i)$ that created or destroyed the singularity \cite{Poulenard}.

\section{Related Work}
\label{sec:related}

Our work examines how \gls{TDA} can be used for \gls{TL}. This section will first examine how our work fits into the body of research on \gls{TL}. Then, we will look at the development of \gls{TDA} tools that can be used to train neural networks.

Previous work in \gls{DA} has focused on aligning feature spaces. The \gls{ADDA} network \cite{tzeng2017adversarial} finds features that maximize a \gls{GAN} style domain loss. As a result, a discriminator cannot distinguish between the source and target in a latent feature space. A classifier is trained on the source representation, and then directly applied to the target representation. The \gls{GAN} style loss has a goal similar to gradient reversal \cite{Ganin15MLR}, but it avoids the issue of vanishing gradients. This overall approach is flexible as it allows different network architectures and training schedules for source and target.  The standard approach is to first initialize a target feature extractor with a source feature extractor after some training on the source data. 

The \gls{DAN} \cite{Long2015} uses \gls{MMD} to align the source and target feature representations. This metric measures the difference between the source and target distribution through the distance in a \gls{RKHS}. The network uses MMD to minimize the discrepancy between the source and target distributions in later task-specific layers of the network. It accomplishes this by matching outputs for source and target. This is problematic because there is not always an ideal correspondence. For example, if you are trying to transfer knowledge from pictures of cats and dogs to pictures of cars and buses it is not apparent how to make a choice in matching. To improve on this design, the \gls{JAN} \cite{Long2017} used a \gls{JMMD} to better minimize the joint discrepancy between the source and target representations not only in task specific layers but also in feature extraction layers.

Recent works by SiSi\cite{SiSi2010}, Mendoza-Schrock \cite{Mendoza2017}, and Rivera et al. \cite{Rivera2020} have employed optimal transport models to align source and target features in a latent space. In general, optimal transport converts one probability distribution into another. 
Our experiments employ the \gls{DiSDAT} architecture \cite{Rivera2020}. This design allows for greater flexibility in the choice of data sets by using separate embeddings for the source and target, then aligning the representations in a common feature space. This network aligns the source and target using \gls{BD} Minimization. \gls{BD} measures the dissimilarity between the source and target distributions in the latent space. It explicitly models the differences between the distributions in the latent space and learns feature mappings such that the source and target distributions match. A main advantage of this divergence measure is that it does not try to match network classifier outputs but instead aligns the feature space distributions that are passed to a network classifier.

Previous network architecture schemes have also utilized separate embeddings for the source and target. In the TLDA network \cite{zhuang2015supervised2}, separate paths are used. However, rather then aligning using a distance measure, in this design the divergence between estimates of the two probability density functions is penalized. Effectively this aligns the sample means of the source and target in the latent space rather than the distributions.

Our approach represents a departure from this framework in that we use tools from \gls{TDA} to make the source and target data manifolds more similar in the latent space. Previous attempts at using \gls{TDA} have focused on clustering and dimensionality reduction.  An algorithm was developed by Niyogi \etal \cite{Niyogi} for estimating the persistent homology in a data manifold created during an unsupervised learning task. In this theoretical work, they show that the homology can be estimated using their algorithm provided a point cloud has a small enough variance. They generalize similar results about multivariate Gaussian distributions, however they do not give a practical application to a real unsupervised learning task.

Although \gls{TDA} in \gls{TL} has been largely unexplored, these tools have been used in deep learning frameworks. The application of TDA to deep learning typically comes in two styles:
 \begin{enumerate}
 \item Kernel methods where a distance between barcodes must be defined.
 \item Feature methods where a real valued vector (a sample image) is associated to each persistence diagram thereby transforming the output of persistent homology into a feature vector suitable for deep learning. 
 \end{enumerate}
 The latter method is also known as preprocessing, and was used by Giansiracusa, Giansiracusa and Moon\cite{GiGi} to improve fingerprint classification. They use a greedy feature selection algorithm to find a featurization vector for classification. Later, Hofer \etal \cite{Hofer} created a trainable input layer based on the persistence diagram. These methods might not be as relevant to \gls{TL} because the feature spaces of the source and target are not necessarily similar. In addition they do not rely on a deep neural networks ability to identify robust features but rather a featurization is chosen a priori.

In order to further the use of \gls{TDA} in \gls{MacL}, Poulenard \cite{Poulenard} developed a differentiable and invertible map from the persistence diagrams back to the individual data points. They observed that the birth and death of a topological feature is the result of a new connection being formed between two data points. Their map was first used by Gabrielsson \etal \cite{SRpointcloud} to reconstruct a surface using point cloud data.
 
Recently, Gabrielsson \etal have brought increased attention to \gls{TDA} for \gls{MacL} by  creating a Pytorch \cite{Pytorch} implementation of the persistence calculation and the invertible map \cite{Gabrielsson2019_topolayer}. They apply a persistent homology based regularization scheme in order to incorporate topological priors in generative networks, and to perform topological adversarial attacks. Using their design, the persistence diagrams of the feature representations can be calculated and differentiated for any layer in the network architecture. Measurements based on these diagrams can be used as loss terms. Although they mentioned using this for a transfer task \cite{Gabrielsson2019_topolayer}(Appendix Section 4), they did not report any \gls{TL} results.

We investigate the use of \gls{TDA} in \gls{TL} tasks. Rather then computing homology for individual images, we consider the entire feature representation of datasets given by a neural network. Thus the individual images are points in a high dimensional space. We therefore examine the persistent homology of the entire data set in aggregate. The key hypothesis is that the global data structure can be used during training to increase accuracy on target data.

\section{METHODS}
\label{sec:methods}

\subsection{Persistence Computations}
Initially our experiments focus on how topology can be used in \gls{TL}. In these instances we were not training to optimize a topological constraint. As such, backpropogation was unneccessary. Gudhi is an open source library of \gls{TDA} that has a python interface. We chose a Rips Complex \cite{Rips, gudhiRipsComplex}, in our preliminary experiments as this algorithm decreased computation time. For these initial experiments, distances were measured using the Wasserstein distance \cite{Wasserstein origins, gudhiPersistenceRepresentations}. The use of Wasserstein distance is discussed in Section \ref{distancecomps}.

Our later experiments required the barcode computation to be differentiable in order to regularize based on topological information.  We used the Pytorch extension implemented by Gabrielsson \etal \cite{Gabrielsson2019_topolayer} to build the filtrations, compute the persistence diagrams and backpropogate. There are several options for filtration algorithms including two for point cloud data. Due to the high-dimensionality of our space, we used the Rips Complex to reduce computation time.

\subsection{Distance Computations}
\label{distancecomps}
A measure of divergence between persistence diagrams is essential for implementation of topological regularization. One can view the persistence diagrams as probability distributions. The Wasserstein distance \cite{Wasserstein origins} is a suitable measure because it takes into account the spatial  realization of both sets along with their density. Let $r$ and $c$ be two distributions, $M$ a matrix of weights, and $U(r,c)$ the collection of all maps from $r$ to $c$. We call $P \in U(r,c)$ a transport plan. Then the 1-Wasserstein distance between $r$ and $c$ is defined as: 

$$ d_m(r,c) := \min\limits_{P \in U(r,c)}  \langle P, M \rangle $$

Unfortunately this is computationally intractable as the number of maps between the two sets increases factorially with the size of the sets. As such, approximation schemes have been used in previous machine learning works \cite{WGAN}. 

In our design we use Sinkhorn iterations\cite{sinkhorn} to simplify the computation. The Sinkhorn approximation to the Wasserstein distance adds a parameter $\alpha$ that can be thought of as an entropy threshold. Instead of looking at all possible maps, the algorithm looks only at plausible maps, where plausibility is measured by entropy. The underlying assumption is that finding maps with low transport cost is more likely with a low entropy transportation plan. We used the Pytorch implementation developed by Daza \cite{sinkhorngit}.

\section{EXPERIMENTS}
\label{sec:experiments} 
In the previous sections, the relevant background material and the state of current research surrounding \gls{TL} and \gls{TDA} are discussed. The following set of experiments attempts to determine how these two bodies of knowledge can inform and improve each other. The goal is to determine best practices for practical application of persistent homology in \gls{TL} tasks.
\subsection{Experiment 1: \gls{TDA} detects object features}
Experiment 1 investigates the amount of topological structure in corner versus center regions of images. We hypothesized that measurements using \gls{TDA} would show that the object-centered pixel information is more valuable than the corner information.

\subsubsection{Experiment 1: Set Up}
To test this, a sample set of 200 images was selected, and Gaussian noise was added. For all 200 images the center and corner $10 \times 10$ pixels were cropped. Each grid was then interpreted as a $100$-dimensional vector, and the $200$ points formed a point cloud in $\mathbb{R}^{100}$. The persistent homology of this point cloud and the lifetimes of the detected topological features were computed. Recall that the lifetime of a topological feature is a measure of how persistent the feature is. Longer lifetimes correspond to more robust topological features, and shorter lifetimes indicate smaller topological features. These small features can be safely ignored since they change depending on the sample set used. Although there may be the same number of holes in two samples of corner data, the actual holes present are located in different places. Hence shorter lifetimes indicate background noise. If the average lifetime of a point cloud generated by the center pixels is longer then that of the corner pixels, this would indicate more topological structure in the object centered regions.

\subsubsection{Experiment 1: Results}
\begin{figure}[ht]
  \begin{centering}
$ \vcenter{\hbox{\begin{overpic}[unit=1mm, scale = .4]{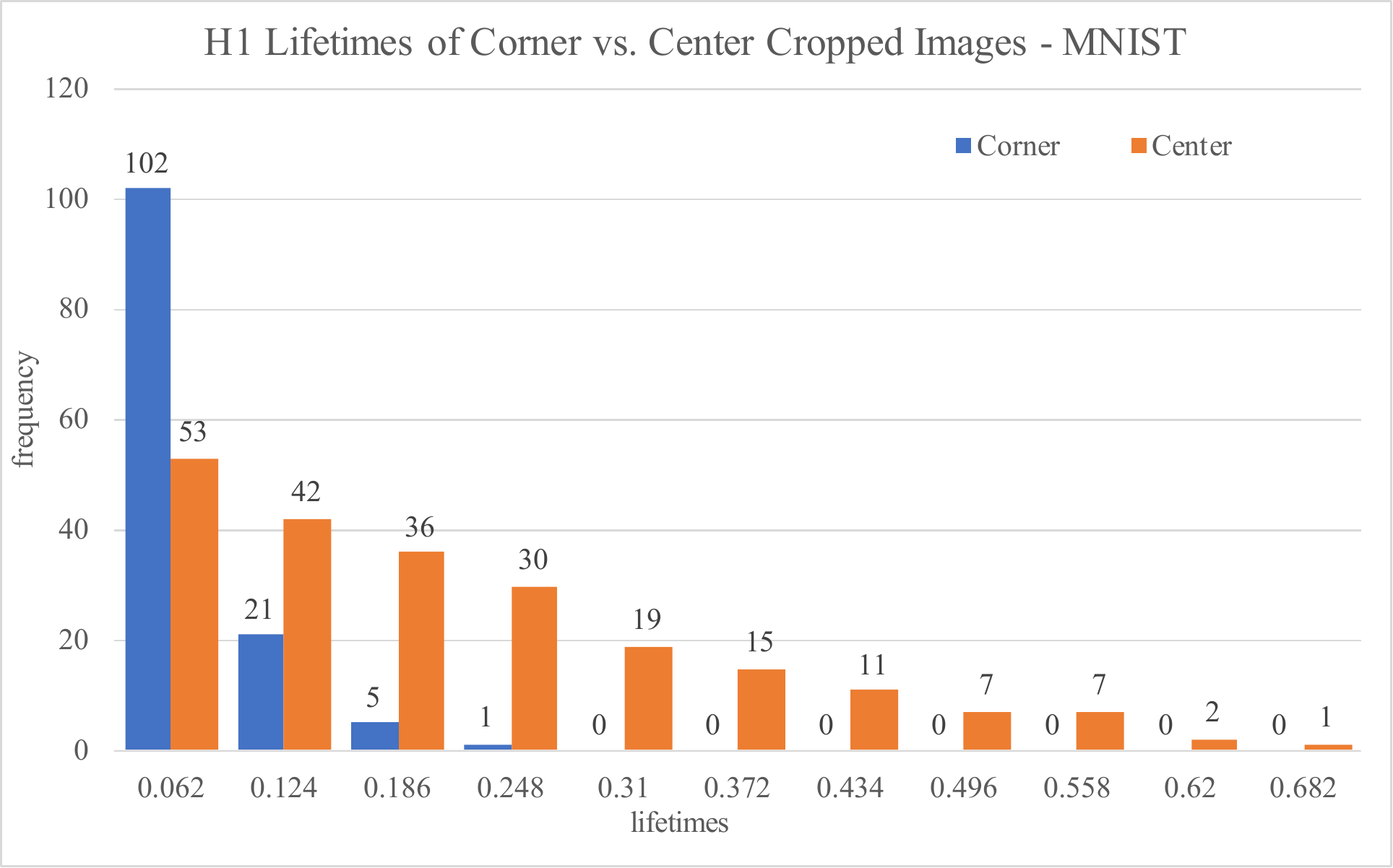}\end{overpic}}}$ 
$\vcenter{\hbox{\begin{overpic}[unit=1mm, scale = .4]{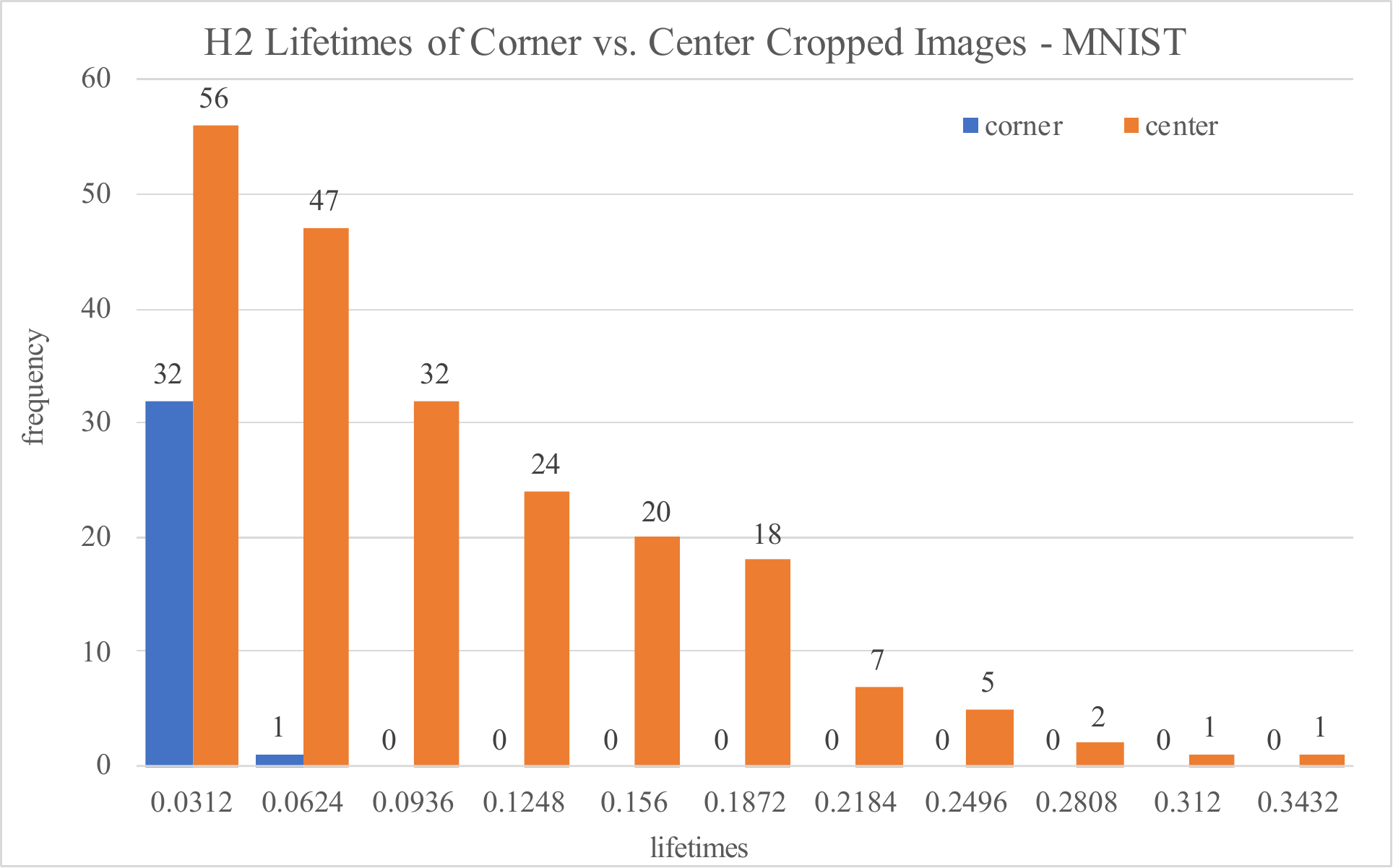}\end{overpic}}}$\\
$ \vcenter{\hbox{\begin{overpic}[unit=1mm, scale = .4]{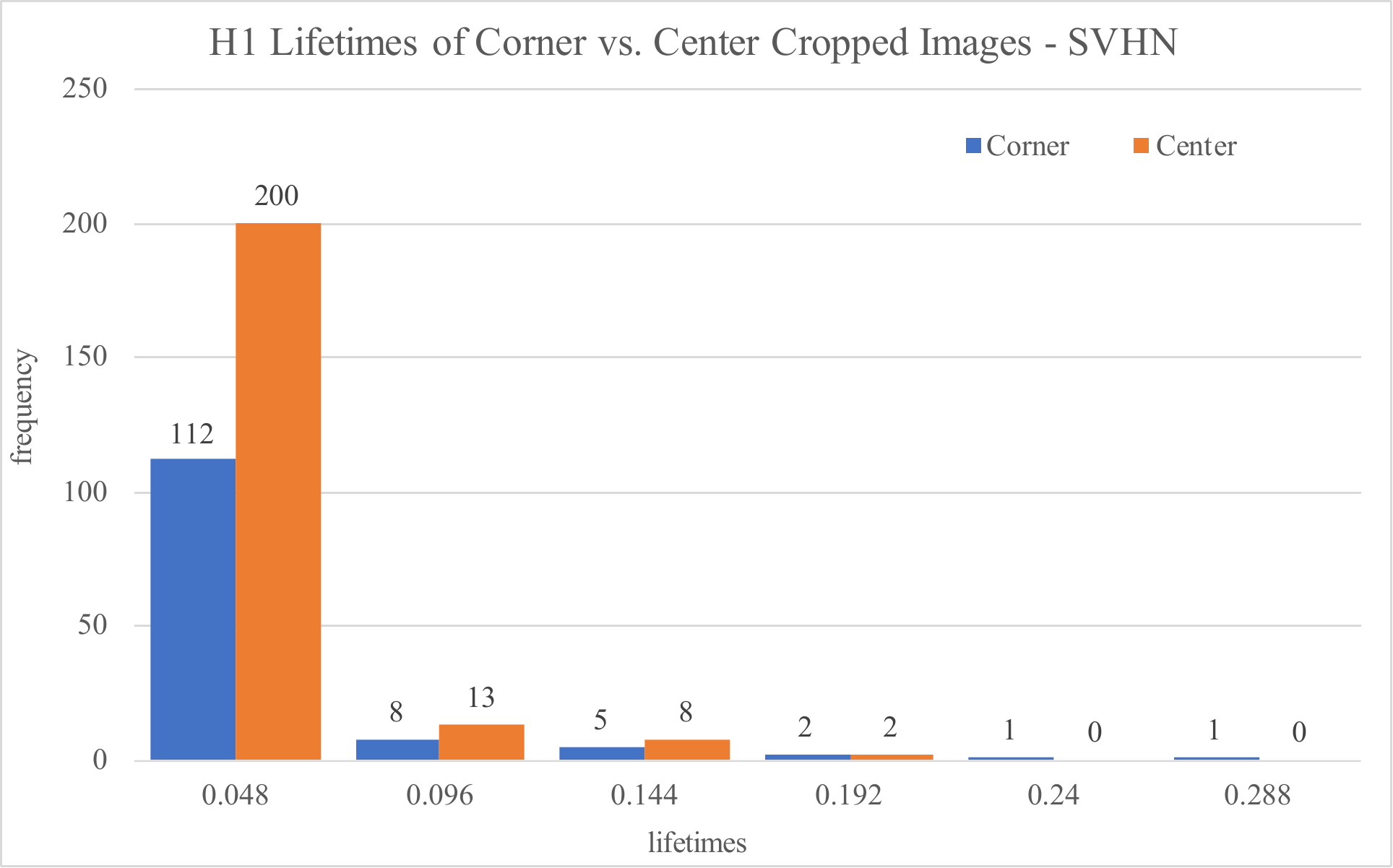}\end{overpic}}}$
$\vcenter{\hbox{\begin{overpic}[unit=1mm, scale = .4]{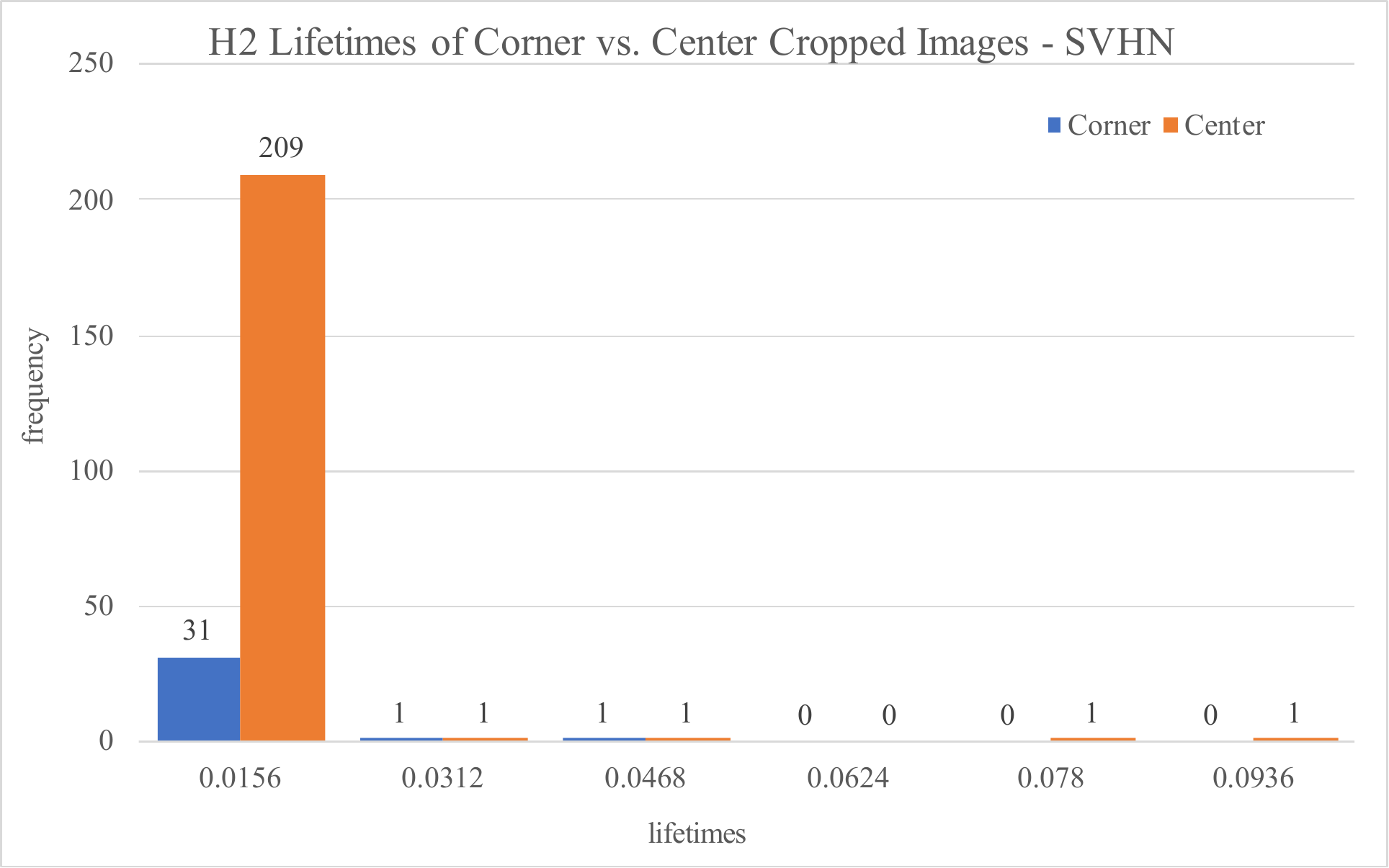}\end{overpic}}}$\\
  \end{centering}
      \caption{The lifetimes of structures are longer for the centered object features then for noisy background features. Background noise has little topological structure.}
      \label{exp2results}
  \end{figure}

As seen in Figure \ref{exp2results}, the center parts of MNIST images generated point clouds with longer-lived topological features then those from the corners. In the top chart for corner samples we see that the lifetime frequency drops off quickly. In the center samples the lifetimes continue to persist for values farther from $0$. There is a less pronounced change for the SVHN data set. Since SVHN is made up of natural images there is more background variability in the corners. Also, the object centered portion of an SVHN image is not always in the center, which is one of the challenges of using this data set in transfer tasks.

 These results indicate that persistence diagrams can be used to detect object features for the MNIST data set. The features pertinent to classification lie in the center of the MNIST images. The level 1 and level 2 persistence diagrams indicate the presence of larger holes in the center manifold then in the corner manifold. This means there are more topological features in the point cloud made by center pixels. The smaller holes in the corner manifold indicate noisy topological features and a lack of large structural components. There is a clear difference in the persistent homology of the corner manifold and the center manifold. This difference can be exploited to distinguish between the parts of the image that contain features used for classification and the parts with background variation. By measuring the average lifetime of one and two dimensional holes and selecting the cropping that gives the longest lifetime, one can select the part of the image with features pertinent to classification and filter out background variation. Since persistent homology could be used to detect features, the next natural question is whether these features are consistent across different samples of the data set.

\subsection{Experiment 2: Persistence Diagram Stability Across Samples}

After establishing that persistent homology can detect features, we investigated whether the persistent homology barcodes were consistent across different subsets of images. Further, we investigated how discontinuous transformations altered the consistency of the barcodes. We tested this by comparing the persistent homology for sets of images where we focus on the object centered pixels to the persistent homology of background noise and pixels that have undergone a discontinuous transformation.

\subsubsection{Experiment 2: Set Up}
To test these assumptions, we used the MNIST and SVHN data sets. First, a random subset of 200 images, called the sample set, was selected from the dataset. Random Gaussian noise was added to this sample set of images. The entire sample set of $200$ images was subjected to one of three conditions - center cropping, corner cropping, and shuffling. In center cropping, the center $10 \times 10$ pixels were selected. For corner cropping, the top left $10 \times 10$ pixels were selected, and for shuffling the center $10 \times 10$ pixels were selected then randomly shuffled with a different permutation applied to each image. 

\begin{figure}[ht]
\begin{centering}
$ \vcenter{\hbox{\begin{overpic}[unit=1mm, scale = .5]{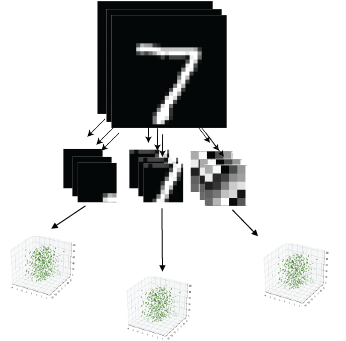}\end{overpic}}}$\\
\end{centering}
\caption{Each subset of 200 images undergoes a transformation creating three separate point clouds. Image from MNIST \cite{MNIST}}
\label{experiment1setup}
\end{figure}

After cropping, each $10 \times 10$ grid was interpreted as a 100-dimensional point and all 200 of these points created a point cloud in $\mathbb{R}^{100}$. See Figure \ref{experiment1setup}. The persistent homologies of these three point clouds were computed. Let $PD_k(Center_1)$, $PD_k(Corner_1)$, and $PD_k(Shuffle_1)$ refer to the persistence diagrams generated by the center, corner, and shuffled images respectively.
Another subset of 200 images was selected, and the process was repeated, forming the persistence diagrams $PD_k(Center_2)$, $PD_k(Corner_2)$, and $PD_k(Shuffle_2)$. 

The 1-Wasserstein distance was then calculated between the persistence diagrams - with $PD_k(Center_1)$ being compared to $PD_k(Center_2)$ - and similarly for the other point clouds. This distance allowed us to quantify the topological similarity of the two samples.

\subsubsection{Experiment 2: Results}
\begin{figure}[ht]
  \begin{centering}
$ \vcenter{\hbox{\begin{overpic}[unit=1mm, scale = .6]{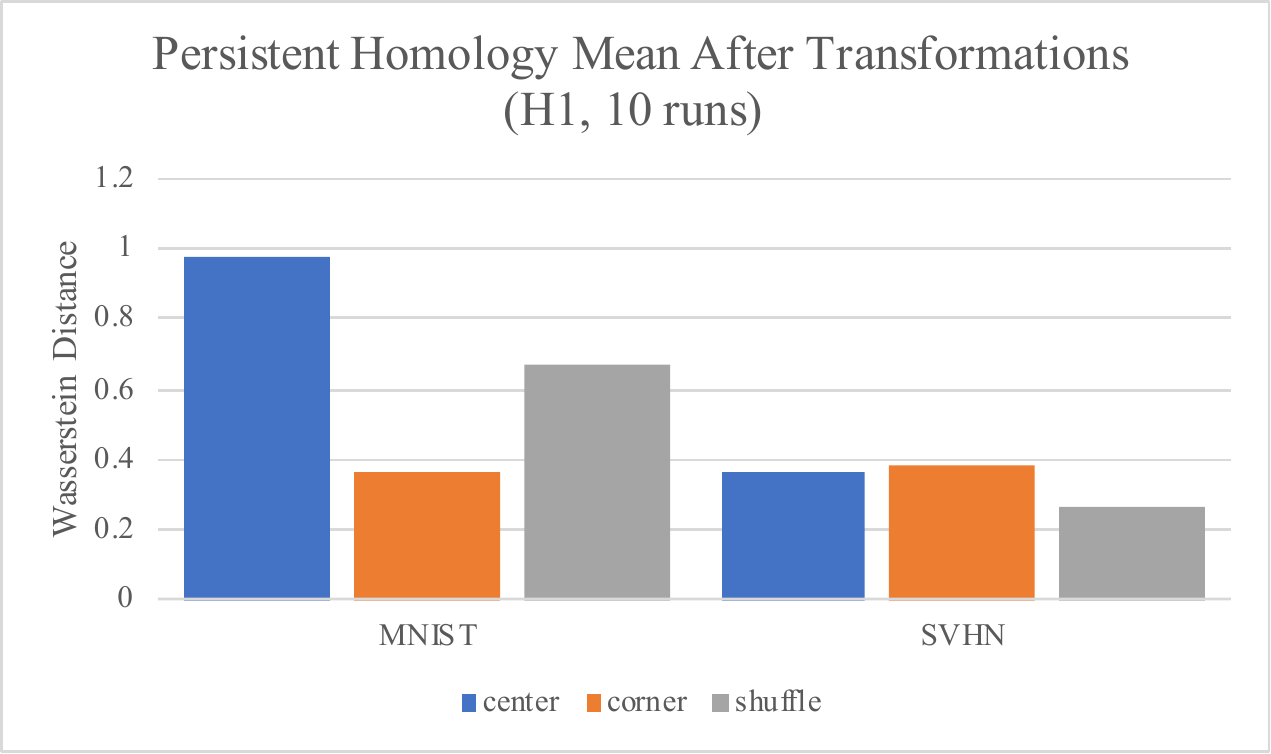}\end{overpic}}}$ \ \ \ \ 
$\vcenter{\hbox{\begin{overpic}[unit=1mm, scale = .6]{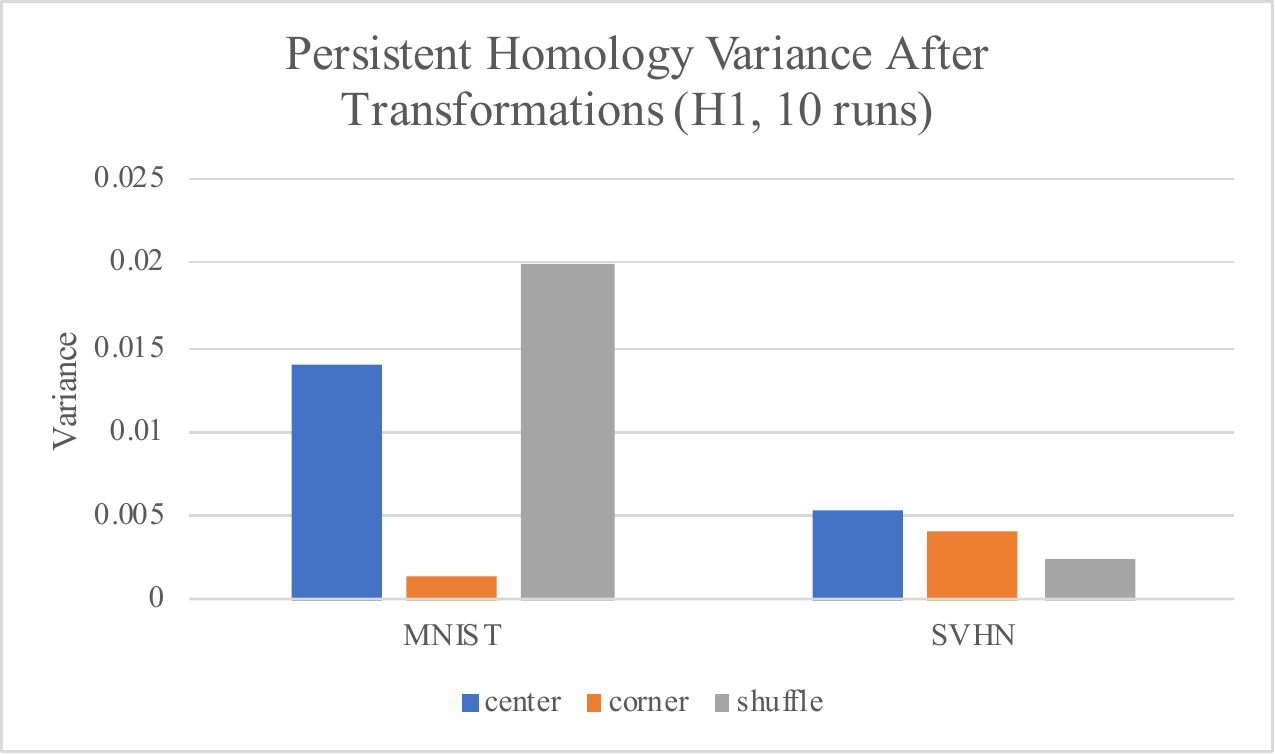}\end{overpic}}}$\\
  \end{centering}
      \caption{Surprisingly, the persistent homology was not consistent for object centered features. Background noise exhibits relatively consistent persistent homology.}
      \label{experiment1results}
  \end{figure}

Figure \ref{experiment1results} gives the results of using the MNIST and the SVHN data sets for generating the image subsets. Surprisingly, the results show that the corners of the MNIST diagrams are more similar across data sets than the object centered portions. The shuffled pixels are also more consistent then the center across samplings, however they are less similar then the corners. The variance indicates even more consistency for the corner cropped subset. For the SVHN dataset, the shuffled pixels were most consistent with center and corner features being roughly equal. 

These results indicate that although topology can determine where the features are within an image, the point clouds generated across different subsets are not consistent. Different subsets create very different topological structures for center cropped images, whereas the subsets generated by corner or shuffled images are consistent. The inconsistency in persistence diagrams could have been produced by simply having more topological structure in those point clouds. Since the corners of MNIST images are mostly black space, the point clouds generated by those portions of the images is a tightly clustered mass of points with little topological structure. This could also explain the discrepancy between the MNIST and SVHN findings. The SVHN data set is made up of naturally occurring images. As such, the background has more topological structure than the background of MNIST samples. This increase in background structure would cause there to be little difference between the topological structure of backgrounds of the SVHN images than that of the object centered regions. Experiment 1 gives evidence to support this since longer lifetimes were found in the center parts of images as opposed to the corner. From these two experiments we know that persistent homology can be used to detect features, although these features may not be consistent across different subsets of images. The next natural question was how current \gls{TL} techniques impact the topology of the source and target representations - the goal being to determine a strategy for \gls{TL} that leverages this tool to improve performance.
 
 \subsection{Experiment 3: The impact of regularization on feature space topology}
 \subsubsection{Experiment 3: Hypothesis}
During training, the network learns a source and target distribution in the latent space. A classifier is learned on the source distribution and then repurposed for the target distribution. During \gls{TL}, different algorithmic approaches are used to align the source and target distributions. We wish to examine what happens to the topological structure of the source and target distributions when these algorithms are applied. The goal was to make an informed decision about how and where to use topological properties in the network. One possibility is that as the network trains robust topological features emerge in the latent space. Since the $0$-level persistent homology quantifies the clustering behavior of a point cloud, it is also possible that as the network trains on a transfer task the source and target may be come more topologically similar.

\subsubsection{Experiment 3: Setup}

The DiSDAT network architecture is a flexible \gls{TL} network designed to accommodate a wide variety of transfer tasks. It features a latent space occupied by both the source and target feature representations where various regularization techniques can be applied. We set up our network, using the general \gls{CNN} architecture of ADDA \cite{tzeng2017adversarial}, but with a latent space dimension of 500, one path for the source and target embedding, and the additional autoencoder of Rivera \etal \cite{Rivera2020} with a parameter of 1.  We trained for five Monte-Carlo iterations with 10 epochs each, and did not pre-train on the source only data. The first conditions corresponds to no regularization. We investigated the topological structures of the latent space under three different regularization conditions. The second condition adds  \gls{DAd} regularization using the \gls{GAN}-style loss. The third condition adds \gls{BD} Minimization to align the source and target probability distributions in the latent space. In this experiment, the source data was the MNIST data set, the target data was the USPS data set.
 \begin{figure}[ht]
  \begin{centering}
$ \vcenter{\hbox{\begin{overpic}[unit=1mm, scale = 1]{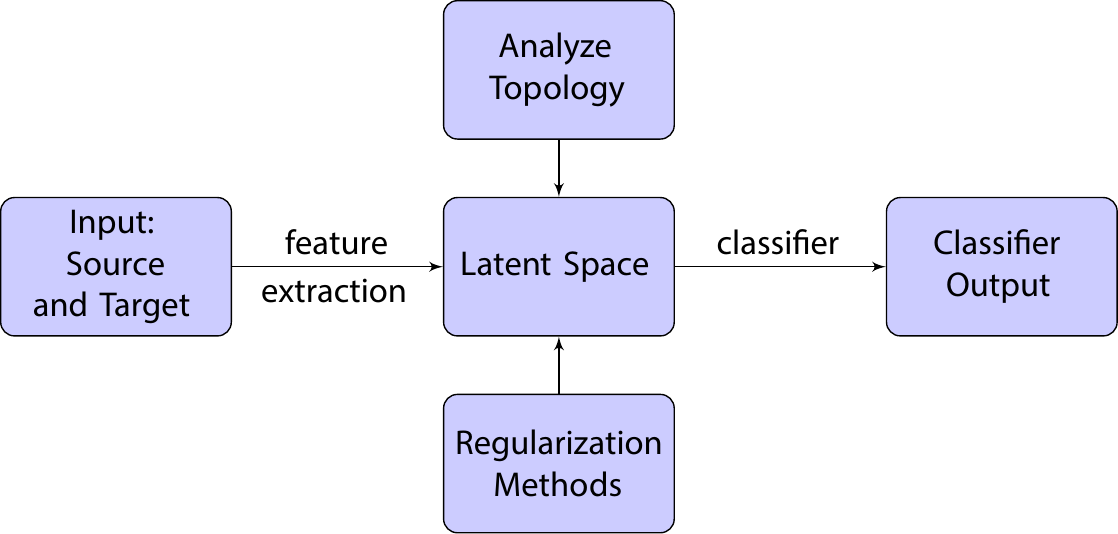}\end{overpic}}}$ 
\\
  \end{centering}
      \caption{The DiSDAT architecture used for the analysis of feature representation topology in \gls{TL} networks.}
      \label{exp3Disdat}
  \end{figure}

At the end of each epoch of training, the last batch was sent through the network to obtain the latent space feature representation of that batch. Then the persistent homology of the source data manifold was computed separately from that of the target data manifold. Once the 0,1, and 2- level persistent homology was computed, the lifetimes of the source and target topological structures were computed. In addition, the Wasserstein - 1 distance between the source and target persistence diagrams was recorded.

\subsubsection{Experiment 3: Results}
\begin{figure}[ht]
  \begin{centering}
$ \vcenter{\hbox{\begin{overpic}[unit=1mm, scale = .4]{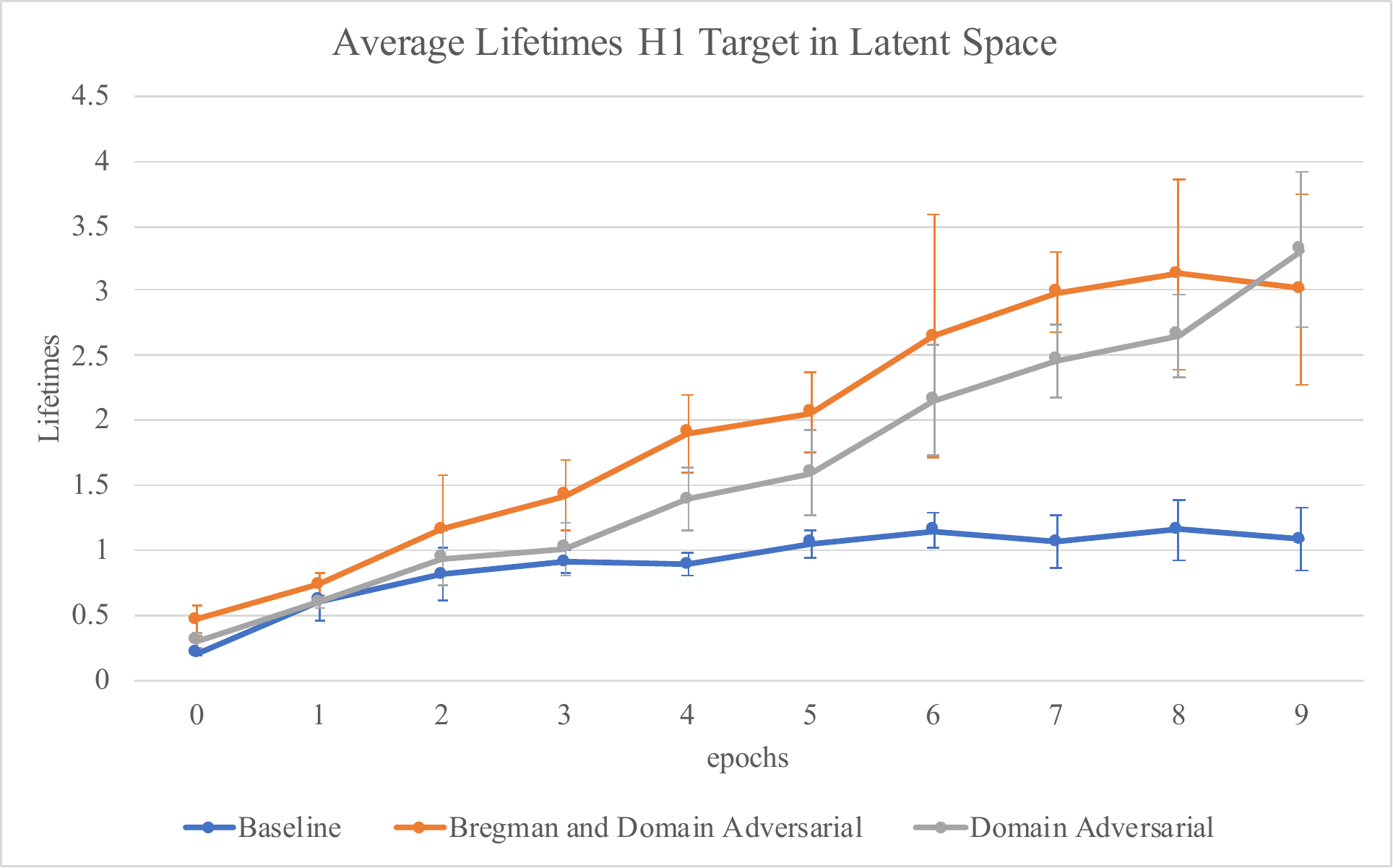}\end{overpic}}}$ 
$\vcenter{\hbox{\begin{overpic}[unit=1mm, scale = .4]{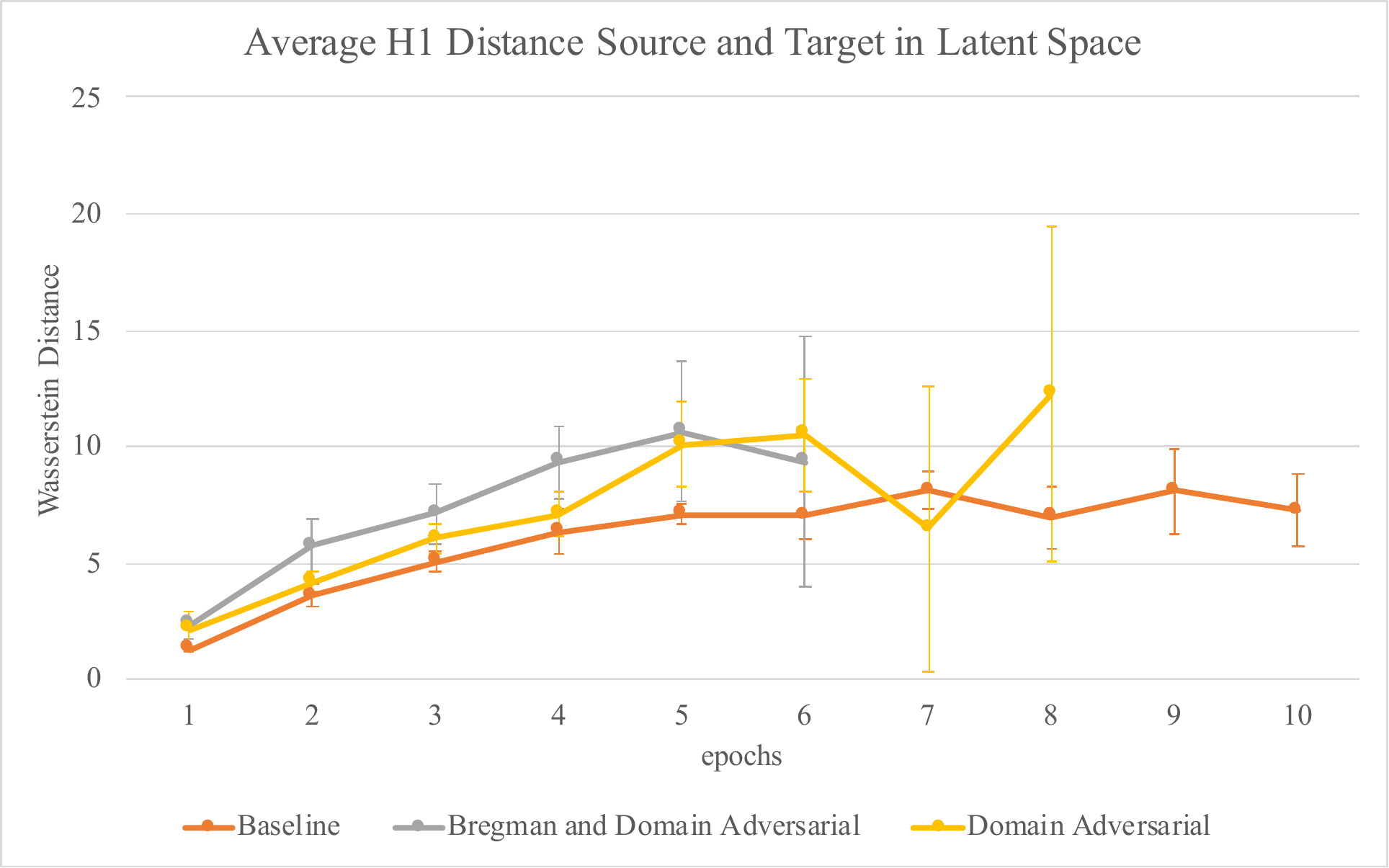}\end{overpic}}}$\\

  \end{centering}
      \caption{Results of experiment 3. In the graph to the right, we see that initially the Wasserstein distance between source and target persistence diagrams grows over training, however eventually it stabilizes. The distance was largest for networks with regularization. In the graph on the left, we see that regularization techniques promote additional topological structure in the latent space.}
      \label{exp3results}
  \end{figure}

The results of experiment 3 are given in the chart in Figure \ref{exp3results}.
  
  We see that the different types of regularization have an impact on the topology of the source and target latent space representations. The chart on the left shows that the addition of regularization increases the lifetimes of the topological structures. This means there are larger topological features with regularization. In the chart on the right, we see that initially the distance between the source and target persistence diagrams starts small, and grows over time. However this growth may actually be the result of having more robust features. As previously discussed, noisy data produces small features but the persistence diagrams of point clouds produced by noise are very consistent. We see that regularization promotes more structured data, and frequently when more structure is present the distances between diagrams can increase. This increase is likely due to the network learning topological features during training rather then a divergence increase between the source and target topologies.
  
\subsection{Experiment 4: Determining the network layers that produce robust topological features}  
\subsubsection{Experiment 4: Hypothesis}
The increase in topological structure in the latent space indicates that the network is learning topological features while training. The ultimate goal of these experiments was to determine how best to implement a topological penalty in a \gls{TL} network. In order to place such a layer in the most advantageous spot, we must determine where in the network architecture topological information is learned. We found that the topological structure in the latent space was more significant then at other layers within the feature extraction section of the network. This led us to apply topological penalties to this layer in Experiment \ref{exp5}. 

\subsubsection{Experiment 4: Setup}

This experiment investigates the lifetimes during training over the various layers. The network was set up in the same way as for the previous experiment, and the same number of epochs and Monte-Carlo iterations were used. The regularization condition used \gls{BD} minimization and  \gls{DAd} regularization. At the end of each epoch of training, the last batch was sent through the network, and the output of each layer was recorded. Then the 0,1, and 2 level persistent homology was calculated for each layers output. The average lifetime was found then recorded. The result was the average lifetime of each layer of the network over training.
\begin{table}[h]
\centering
\begin{tabular}{||l || l ||} 
 \hline
  Layer & Type \\ [0.5ex] 
 \hline\hline
1 & Convolution (2d, input channels = 1, output channels = 20, kernel size = 5) \\ 
2 & Max Pooling (2d, kernel size = 2) \\
3 & ReLU activation \\
4 & Convolution (2d, input channels = 20, output channels = 50, kernel size = 5 \\
5 & Drop Out (2d)\\
6 & Max Pooling (2d, kernel size = 2)\\
7 & ReLU activation\\
 [1ex] 
 \hline
\end{tabular}
\caption{Experiment 4 - network implementation details for the LeNet encoder. The output of layer 7 is what is referred to as the latent space.}
\label{Experiment4table}
\end{table}

\begin{table}[h]
\centering
\begin{tabular}{||l || l ||} 
 \hline
  Layer & Type \\ [0.5ex] 
 \hline\hline
1 & Drop out (1d, parameter = 0.5) \\ 
2 & Linear (input = 500, output = 500)\\
3 & ReLU activation \\
4 & Drop out (1d, parameter = 0.5) \\
5 & Linear (input = 500, output = 500)\\
6 & ReLU activation\\
7 & Linear (input = 500, output = 784)\\
7 & Tanh\\
 [1ex] 
 \hline
\end{tabular}
\caption{Experiment 4 - network implementation details for the decoder.}
\label{Experiment4tabledecode}
\end{table}

\begin{table}[h]
\centering
\begin{tabular}{||l || l ||} 
 \hline
  Layer & Type \\ [0.5ex] 
 \hline\hline
1 & ReLU activation \\ 
2 & Dropout \\
3 & Linear (input = 500, output = number of classes (10)) \\
 [1ex] 
 \hline
\end{tabular}
\caption{Experiment 4 - network implementation details for the LeNet classifier.}
\label{Experiment4tableclassifier}
\end{table}

\begin{table}[h]
\centering
\begin{tabular}{||l || l ||} 
 \hline
  Layer & Type \\ [0.5ex] 
 \hline\hline
1 & Linear (input = 500, output = 500) \\ 
2 & ReLU activation \\
3 & Linear (input = 500, output = 500) \\
4 & ReLU activation\\
5 & Linear (input = 500, output = 1)\\
6 & Sigmoid\\
 [1ex] 
 \hline
\end{tabular}
\caption{Experiment 4 - network implementation details for the Discriminator}
\label{Experiment4tablediscriminator}
\end{table}

\subsubsection{Experiment 4: Results}
The chart below summarizes the results of experiment 4 run on the DiSDAT network with \gls{BD} minimization and  \gls{DAd} regularization. We see that there is a trend for increased lifetimes during training, with later layers exhibiting more topological structure and more variance between runs. 

     \begin{figure}[ht]
  \begin{centering}
$ \vcenter{\hbox{\begin{overpic}[unit=1mm, scale = .4]{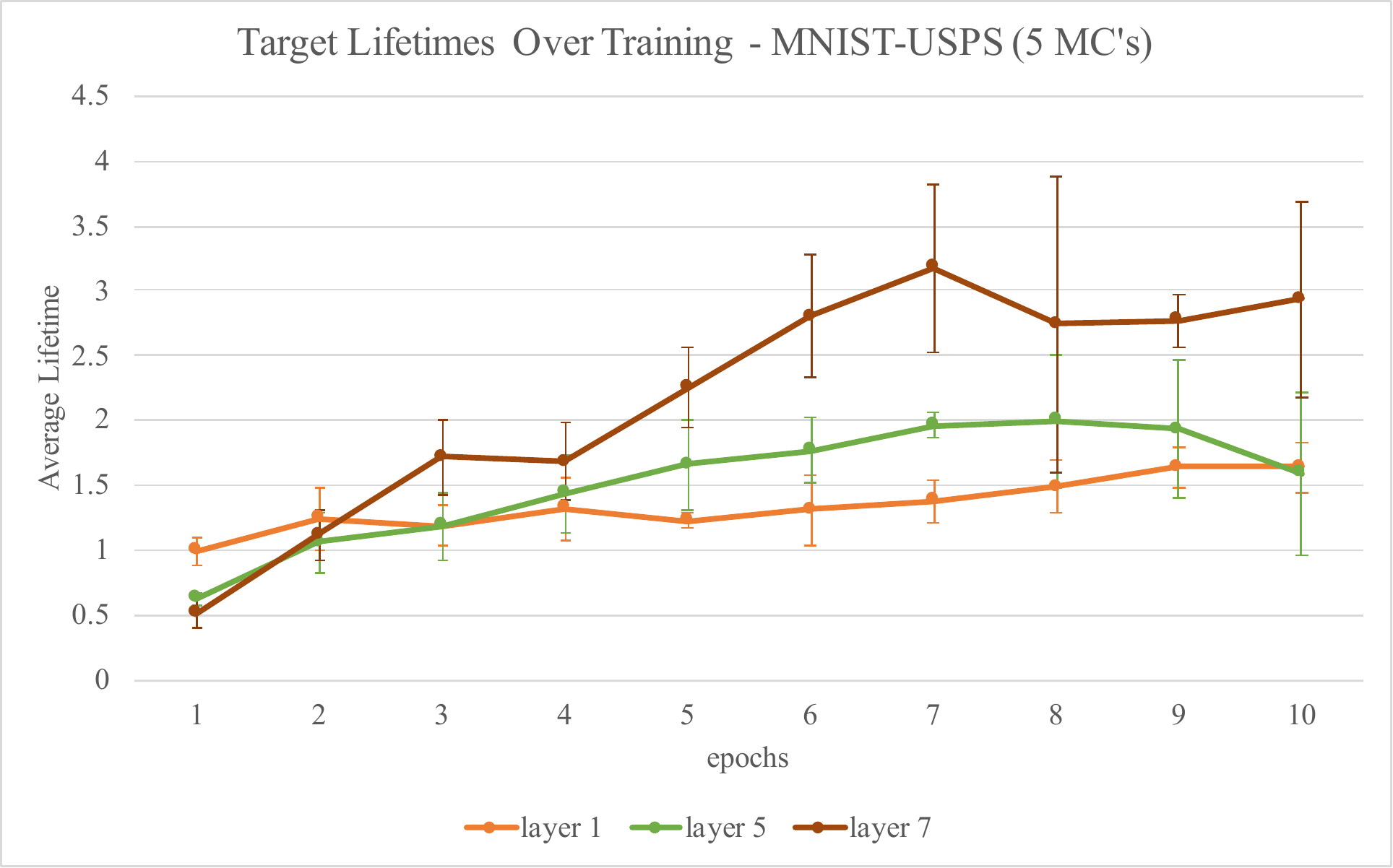}\end{overpic}}}$

  \end{centering}
      \caption{Experiment 4 results for condition with \gls{BD} minimization and  \gls{DAd} regularization. Layer seven has a significant increase in lifetimes of features. Lifetimes grow over training epochs.}
      \label{exp4results}
  \end{figure}
  
\subsection{Experiment 5: \gls{TDA} for \gls{TL}}  
\subsubsection{Experiment 5: Hypothesis}
In summary, our experiments so far suggest that topology alone is not enough to align the source and target in the latent pace. Regularizing with topology may make the data appear noisier since noise produces relatively consistent structure. At most, it can be used to fine tune the network in later stages of training once the topological structure has already emerged. These discoveries prompted a design of a \gls{TL} network that incorporated topological loss. By applying a topological loss to the latent space of the DiSDAT network we hope to force the source and target to have more similar persistence diagrams and thus more similar topological structure in their feature representations. This regularization should have an impact on the performance of the neural network on the MNIST-USPS transfer task.

\subsubsection{Experiment 5: Setup}
\label{exp5}
We begin with a brief discussion of the design of the DiSDAT network \cite{Rivera2020}. The network learns using a sequence of parameter updates. With each batch, the network first computes the  autoencoder loss and accuracy loss.
The embedding map $\mathcal{F}: X \rightarrow \mathbb{R}^d$ is established using an autoencoder to ensure that the map preserves the geometric configuration of the space while compressing the dimensionality - $d$. Let $\mathcal{H}$ denote the low dimensional latent space. The network projects the feature space into the low dimensional manifold, then reconstructs the data by learning a decoding map, $\mathcal{D}:\mathcal{H} \rightarrow X$, back to the original feature space. The Mean Square Error is used to calculate the distance between the original input vector, and the output of the decoder. Therefore we have:
$$L_{ae}:=L_{MSE}(x^s, \mathcal{D}(\mathcal{F}(x^s)))+L_{MSE}(x^t, \mathcal{D}(\mathcal{F}(x^t)))$$
This loss is computed, and the network weights are updated using backpropogation.

Now we turn our attention to the accuracy loss. A classifier $\mathcal{C}: \mathcal{H} \rightarrow Y$ must be learned that minimizes the classification error with respect to some measure after the autoencoder has mapped the data to the latent space. We use cross entropy loss:
\begin{align*}
L_c :=E_{(x,y) \sim P^s} - \langle (y, log(\mathcal{C}(\mathcal{F}(x))))\rangle
\end{align*}
Here, y is a one-hot vector which is zero except for a one at the index of the correct class, and $\mathcal{C}(x)$ is the classifier output. This loss is computed, then the network updates to minimize this loss as before.

Next, the network uses domain confusion as a divergence penalty. We do not use gradient reversal \cite{Ganin15}, and instead opt for a GAN style loss because this method is less susceptible to vanishing gradients \cite{tzeng2017adversarial}. This approach to optimization makes our autoencoder similar to the adversarial autoencoder method in Makhzani \etal \cite{Makhzani2016}. The network optimization iterates between two adversarial criteria. First, a domain classifier's parameters are updated to decrease the domain classification binary cross entropy loss. Next the feature extractor parameters are optimized to increase the domain classification loss. During the iterative optimization, each step is applied in turn. Formally, let $\mathcal{C}_D:\mathcal{H} \rightarrow {0,1}$ be the domain classifier that distinguishes between source and target data. Domain confusion loss, $L_d$, is therefore:
\begin{align*}
L_d := E_{x \sim P^s}(\ln \mathcal{C}_D(\mathcal{F}(x))+E_{x \sim P^t}(\ln(1-\mathcal{D}_D(\mathcal{F}(x))))
\end{align*}
The loss is computed, and the network iteratively updates to minimize the loss as described above.

Next, the probability distributions of the source and target are aligned using Bregman divergence minimization. We use a divergence penalty following Rivera \etal \cite{Rivera2020} 

Last, the network trains based on topological loss, attempting to minimize the Wasserstein distance between the source and target persistence diagrams. See Section \ref{distancecomps} for more information on this loss function and how it was approximated.
\begin{align*}
L_{top}:= Wass(P^s, P^t)
\end{align*}
This is minimized to force the topology of the source and target to be as similar as possible. Implementation of this requires backpropogation through persistence diagrams. This process is discussed in detail in the Section \ref{sec:related} and Section \ref{sec:Background}. This is the final loss in our cascade. After computing this loss, the network updates one last time before proceeding to the next batch of sample points.

We ran the MNIST-USPS task with the following conditions. We opted to initialize the network by running for 30 epochs only on labeled source data. After, we ran for 50 epochs on both the source and target data. An additional 3 epochs of training were then run with the addition of topological regularization in the latent space. The set up was run for 5 Monte-Carlo iterations. There were 64 samples per batch. We ran the experiment on three different network conditions - one with only \gls{BD} Minimization, one with  \gls{DAd} regularization and an autoencoder, and one with \gls{BD} Minimization,  \gls{DAd} regularization, and an autoencoder. 





\subsubsection{Experiment 5: Results}
Figure \ref{exp5results} and Table \ref{Experiment5} below summarize the results of Experiment 5, where topological regularization was added to the loss function. We see that adding topological regularization did not increase the accuracy of the network on target data classification. When only \gls{BD} was used to align the feature spaces, the addition of topological loss harms the accuracy on target data. In the other two conditions, the additional constraint has limited impact. 

We note that given the computation time necessary to regularize with respect to topology, it is unlikely that using topology for more training epochs would result in performance increases significant enough to justify the cost.

\begin{table}[h]
\centering
\begin{tabular}{||c || c c c||} 
 \hline
   & No Topology Loss & $H_0$-loss & $(H_0 + H_1)$-loss \\ [0.5ex] 
 \hline\hline
\gls{BD} & $0.693 \pm 0.0557$ & $0.539 \pm 0.121$ & $0.608 \pm 0.096$  \\ 
  \gls{DAd}  & $0.942 \pm0.0050$ & $0.9308 \pm 0.0038$ & $0.9316 \pm 0.0059$ \\
  \gls{DAd}, Autoencoder,  & $0.9422 \pm 0.0052$ & $0.9321 \pm 0.0066$ & $0.9332 \pm 0.0054$ \\
 Bregman Divergence & & & \\
 [1ex] 
 \hline
\end{tabular}
\caption{Experiment 5 results for three conditions.}
\label{Experiment5}
\end{table}
     \begin{figure}[ht]
  \begin{centering}
$ \vcenter{\hbox{\begin{overpic}[unit=1mm, scale = .5]{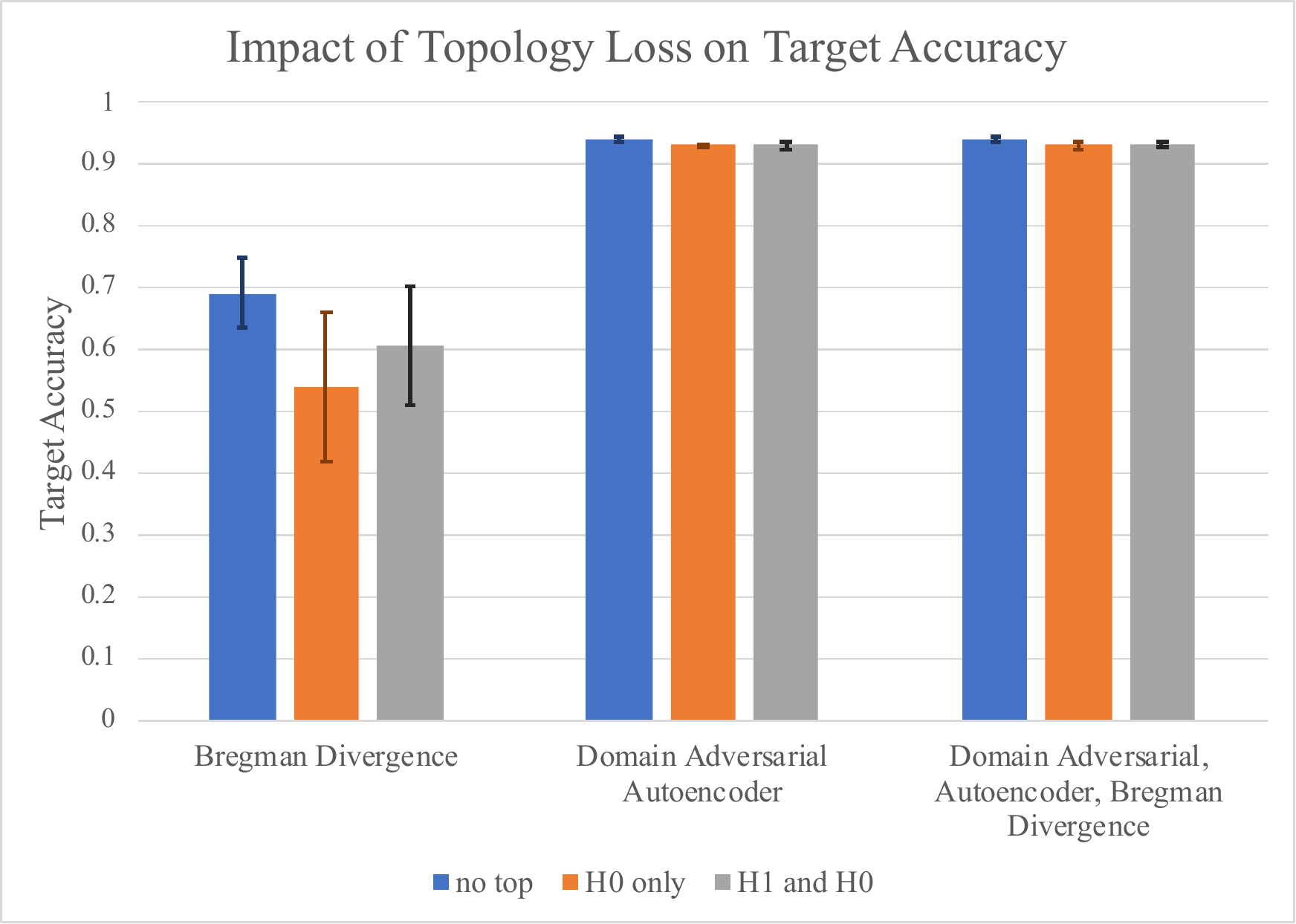}\end{overpic}}}$

  \end{centering}
      \caption{Experiment 5 results for three conditions. Decreased performance was observed in the network with only \gls{BD} minimization. The  \gls{DAd} Autoencoder condition, along with the  \gls{DAd} autoencoder \gls{BD} condition, both showed approximately the same accuracy with and without topological loss.}
      \label{exp5results}
  \end{figure}

\section{Discussion}
\label{sec:discussion} 
The results of these experiments indicate that great care must be taken when attempting to regularize based on topological priors. The results of experiments one through four together indicate that the topological structure of the source and target feature representations are both increasing over training. The increase in distance between the source and target diagrams indicated in experiment three is probably the result of the increased structure. The results of experiment 5 show that forcing the source and target to have topologically similar latent space representations has limited impact on the target classification accuracy. 

Although experiment five does not show increased performance on the transfer task, there could be several causes for this result. First, the algorithm we used to compute the filtration for the homology calculation works for small and medium sized spaces. In a high dimensional space, a batch sample of 64 may be too sparse for the algorithm to connect nearest neighbors between points. Redesigning the network architecture to feature a lower dimensional latent space, or applying the topological regularization to a low dimensional classifier layer may allow the algorithm to find the connections necessary to draw an appropriate simplicial complex.  Another possibility is that the network minimizes the distance between the source and target persistence diagrams by simply making the representations noisier, and thus harder to classify. 

Rather then regularizing to force the source and target to appear more topologically similar, it may be better to use topology to favor good structure within the data set. This is possible by using different featurizations of the persistence diagrams already created\cite{Gabrielsson2019_topolayer}. In our design, we add the divergence between the $H_0$ and $H_1$ persistence diagrams together in the loss function. However, it might make more sense to minimize the topological divergence between source and target for $H_0$ while improving the overall structure present in $H_1$. The  $1$-level homology gives the number of holes and tunnels in the manifold. Finding decision boundaries is easier for the classifier when the feature representations form tight clusters. It might be best to minimize the number of the holes and tunnels by using a different loss metric on the $H_1$ diagrams. Consider the following featurization from \cite{Gabrielsson2019_topolayer} of a $1$-level persistence diagram:

$$\mathcal{E}(p,q,i_0, PD_1) = \sum \limits_{i=i_0}^{I_1} |d_i-b_i|^p \left ( \frac{d_i+b_i}{2} \right)^q $$

By varying the parameters $p, q,$ and $i_0$ you can favor specific conditions related to the number and size of holes. For example, increasing $p$ penalizes persistent features, whereas $q$ favors features that appear later in the filtration. Choosing $p=2$, $q=0$ and $i_0 = 1$ will minimize the number of these structures.

\section{Conclusion}
\label{sec:conclusion}  

This article explores the the use of persistent homology for transfer learning. In this work we treated individual images as points in a higher dimensional feature space, and analyzed the manifold formed by point clouds of images in that feature space. We investigated: 1) how persistent homology can be applied to identifying robust features, 2) the effect of \gls{DA} regularization on persistent homology, 3) the change in homology through network layers over learning, and 4) persistent homology regularization for \gls{DA}.  

Experiment 1 found that \emph{object features} typically exhibit \emph{longer lifetimes} than non-object  image variation, suggesting that longer lifetimes are indicative of a desirable manifold structure. However, experiment 2 demonstrated that the persistence diagrams for object features are not consistent across different image point cloud sets. This means that while persistence can be used to identify good features within sets of images, it is not as helpful for matching similar features across datasets. 

Experiment 3 showed that existing \gls{DA} approaches that use \gls{DAd} or optimal transport regularization result in feature representations with longer lifetimes over a baseline without regularization. This suggests that \gls{DA} may inadvertently produce desirable topological structure without explicitly enforcing topological regularization. Furthermore, experiment 4 showed that the structure improves for later network feature extraction layers. 

Unfortunately, experiment 5 did not find topology regularization to improve transfer performance above existing approaches. We discussed possibilities for this including limited sample size and performance ceiling issues that would be possible avenues for future work. Overall, we have provided several meaningful empirical findings on topology for classification and \gls{TL}.


\acknowledgments     

We would like to thank Dr. Olga Mendoza-Schrock and Mr. Christopher Menart for their input and feedback during this research project. This work was supported by Air Force Research Laboratory (AFRL), Air Force Office of Scientific Research (AFOSR), Dynamic Data Driven Application Systems (DDDAS) Program, Autonomy Technology Research Center (ATRC), and Dr. Erik Blasch.

%


\bibliographystyle{spiebib}   
\bibliography{rivera,weeks}   

\end{document}